# Multi-Attribute Enhancement Network for Person Search

Lequan Chen[1, 2], Wei Xie[1, 2 *], Zhigang Tu[3], Jinglei Guo[1, 2], Yaping Tao[1, 2], Xinming Wang[1, 2]
[1] *Hubei Provincial Key Laboratory of Artificial Intelligence and Smart Learning,*
*Central China Normal University, Wuhan, China*
[2] *School of Computer, Central China Normal University, Wuhan, China*
[3] *State Key Laboratory of Information Engineering in Surveying, Mapping and Remote sensing (LIESMARS),*
*Wuhan University, Wuhan, China*
* Corresponding author: xw@mail.ccnu.edu.cn

*Abstract*—Person Search is designed to jointly solve the problems of Person Detection and Person Re-identification (Re-ID), in which the target person will be located in a large number of uncut images. Over the past few years, Person Search based on deep learning has made great progress. Visual character attributes play a key role in retrieving the query person, which has been explored in Re-ID but has been ignored in Person Search. So, we introduce attribute learning into the model, allowing the use of attribute features for retrieval task. Specifically, we propose a simple and effective model called Multi-Attribute Enhancement (MAE) which introduces attribute tags to learn local features. In addition to learning the global representation of pedestrians, it also learns the local representation, and combines the two aspects to learn robust features to promote the search performance. Additionally, we verify the effectiveness of our module on the existing benchmark dataset, CUHK-SYSU and PRW. Ultimately, our model achieves state-of-the-art among end-to-end methods, especially reaching 91.8% of mAP and 93.0% of rank-1 on CUHK-SYSU. Codes and models are available at https:// github. com/ chenlq123/ MAE.

## I. Introduction

Person Re-identification (Re-ID) is to match the target person in the gallery where the body of the person has been cropped. Person Search aims to find a query person from the whole scene gallery shot by multiple cameras, i.e. it is to jointly solve the two tasks of Person Detection and Re-ID.

Person Search is closer to practical applications than Re-ID, such as pedestrian tracking, video surveillance, etc. Therefore, it is a research topic with broad application prospects in computer vision, and is more challenging. However, Re-ID encounters challenges, e.g. changes in viewpoint of camera, different resolutions, and illumination, etc. In addition to the above, Person Search also needs to face false detection, misalignments in the stage of pedestrian detection. Meanwhile, pedestrian detection focuses on the commonality between pedestrians, while Re-ID focuses on the differences between pedestrian characteristics. The combination of these problems further damages the performance of Person Search.

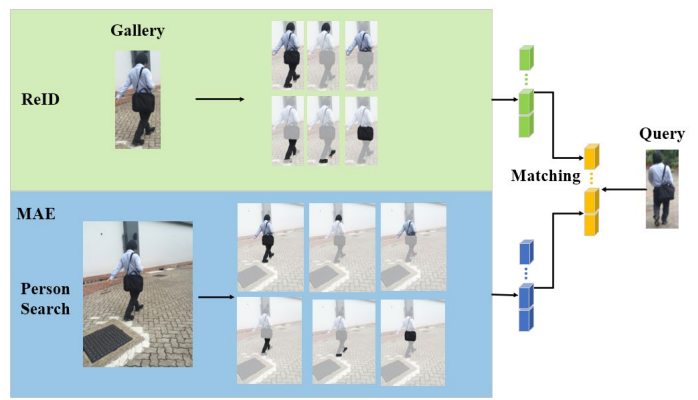

Fig. 1. Comparison attribute alignment in Person Re-identification and Person Search. The current method mainly uses global features for matching or local alignment through horizontal division, while our MAE model learning attribute features are used for matching tasks.

Following Xu et al. [6] in 2014, many methods [8, 13, 20] have been proposed to solve Person Search and achieve important breakthroughs. A classic solution is to utilize a pedestrian detector to obtain bounding boxes, and then Re-ID to find the query person. Most methods [9, 14, 16] use complete pedestrian features for matching. It is worth noting that these features include not only foreground information that is conducive to discrimination, but also background information that damages search performance. Recently, some studies [2, 5] used masks to suppress background information to enhance feature representation. In [8], the Siamese network was designed to extract more refined representations from person patches, thereby avoiding the interference of redundant context information outside the bounding boxes. However, if the background information is simply suppressed, some key attributes related to pedestrians will be lost, such as backpacks, suitcases, companions, etc.

In the Re-ID community, relevant research [29, 30] has pointed out that distinguishing representations play a core role in both the overall and local aspects. Reference [27] use the key points of pose to learn local representations related to body. Reference [38] directly use semantic segmentation to mark body parts for local alignment. However, these methods have a fatal flaw, as shown in Fig. 1, i.e. they are based on the gallery of cropped image. When directly applied to Person Search, there will be more interference clutter in the raw video frames, which will cause serious segmentation errors, and then lead to a linear decline in search performance. How to combine the two effectively is still a long way to go, which requires not only novel model design, but also the support of related datasets.

Inspired by this, in order to make full use of the attributes of pedestrians, as much as possible to enhance the pedestrian representation to distinguish different pedestrians in the recognition stage, we propose a novel multi-attribute enhanced model. In the stage of recognition, in addition to using the global characteristics of pedestrians for matching, it also makes full use of segmentation labels to learn local attribute representation. Specifically, we use segmentation labels to force the model to learn local attribute features from feature maps. We design the attribute learning module, which is responsible for learning the representation of local attributes. Moreover, we notice that the bounding boxes generated by the detector will cause false alarms and misalignments. As shown in Fig. 2, the bounding boxes will not be completely aligned with the ground-truth. Meanwhile, if pedestrian segmentation is performed on uncropped images or bounding boxes, the amount of model parameters and calculations will increase rapidly, and the model inference speed will drop to an unacceptable level. Given these two points, we construct a whole scene segmentation labels suitable for Person Search to achieve the alignment learning and obtain fine-grained features. As shown in Fig. 2, The actual attribute label marked in green is not only aligned with the bounding box, but also easy to train the model.

All in all, the main contributions of this article are three-fold:

- Currently, we are the first to establish whole scene multi-attribute label dataset related to CUHK-SYSU and PRW, which provides more possibilities for introducing multi-attribute matching in Person Search.

- We propose a Multi-Attribute Enhanced model which incorporates attribute features into person representation to enhance the discrimination of features. Experimental results show that this method can improve the performance of detection and recognition.

- Our method is simple and efficient, and achieves state-of-the-art among end-to-end methods on standard benchmarks.

## II. RELATED WORK

In this section, we first introduce person search, and then we review person Re-ID.

**Person Search**：Person Search aims to locate target person in a gallery of video frames. Recently, a large number of methods [12, 14, 15, 17, 20] have been proposed and achieved remarkable results. It is a joint solution to the two tasks of detection and re-identification. Therefore, search performance is restricted by both. According to training method, the current methods can be divided into end-to-end and two-step methods.

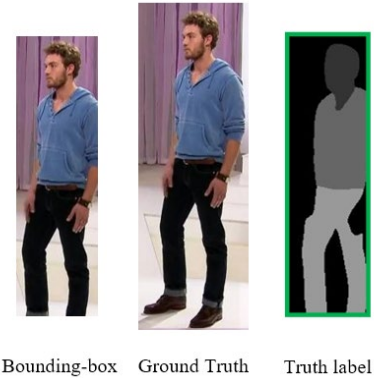

Fig. 2. How to use attribute labels? The first column is the Bounding-box. The second column is the corresponding Ground Truth. The third column is the actual attribute label.

End-to-end methods are friendly, efficient and easy to train. The first is the Online Instance Matching (OIM) proposed by Xiao et al. [14]. Munjal et al. [15] applied a query-guided method to obtain global context from both query and gallery images. Han et al. [16] developed the ROI transformation layer in order to generate more suitable Re-ID bounding boxes so that the proposals could be supervised by re-identifier. Chen et al. [9] demonstrated Norm-Aware Embedding (NAE) decomposes features into radial norm and angle in the polar coordinate, to alleviate the objective contradiction between detector and re-identifier. Zhong et al. [13] proposed an Align-to-Part Network to align the discriminative part features that can be extracted to solve misalignment of bounding boxes.

The two-step methods can alleviate the optimization contradiction between the detector and the re-identifier, so it can obtain better performance. Chen et al. [2] proved this through experiments for the first time. Zheng et al. [20] combined different detectors and re-identifiers and compared the search performance comprehensively. In order to improve the consistency of the two stages, Wang et al. [12] incorporated query similarity calculations into the detector and constructed a hybrid dataset in the stage of Re-ID to improve robustness. Zheng et al. [11] used Siamese network to integrate information of the query into the detection network, and reduced the number of proposals based on the similarity between the target and the proposals.

In this paper, our method is implemented by an end-to-end method based on NAE [9]. We propose multi-attribute enhanced module based on whole scene multi-attribute label dataset. It is worth noting that our labels are aligned with the bounding boxes. Furthermore, the module is also applied to OIM [14].

**Person Re-ID:** At present, methods based on deep CNNs to solve this problem have become the consensus of researchers. Ye et al. [19] pointed out that these methods focus on representation learning [21], metric learning [22] and ranking optimization [23]. Recently, it has become a trend to use attention mechanism to model global and local features as pedestrian representation [27, 28 36, 37]. Chen et al. [24] pointed out that local cues are masked by the most salient features, hence they proposed salience-guided network could adaptively mine potential information of different important levels. Zhu et al. [25] designed the cascade network learning based on feature map, which can not only learn parts of human body, but also learn local features of personal items. We are inspired by this approach, and we also use a concise mask to learn local features and reduce the complexity of the model. The difference is that we did not directly integrate the pedestrian segmentation technology into Person Search, because it will cause a rapid increase in the number of model parameters, resulting in a decrease in running speed. We will deal with these two steps independently. First, the existing pedestrian segmentation technologies are used to obtain attribute labels by semantic segmentation of the people in the video frame, and then the labels are used to supervise and train the model of Person Search.

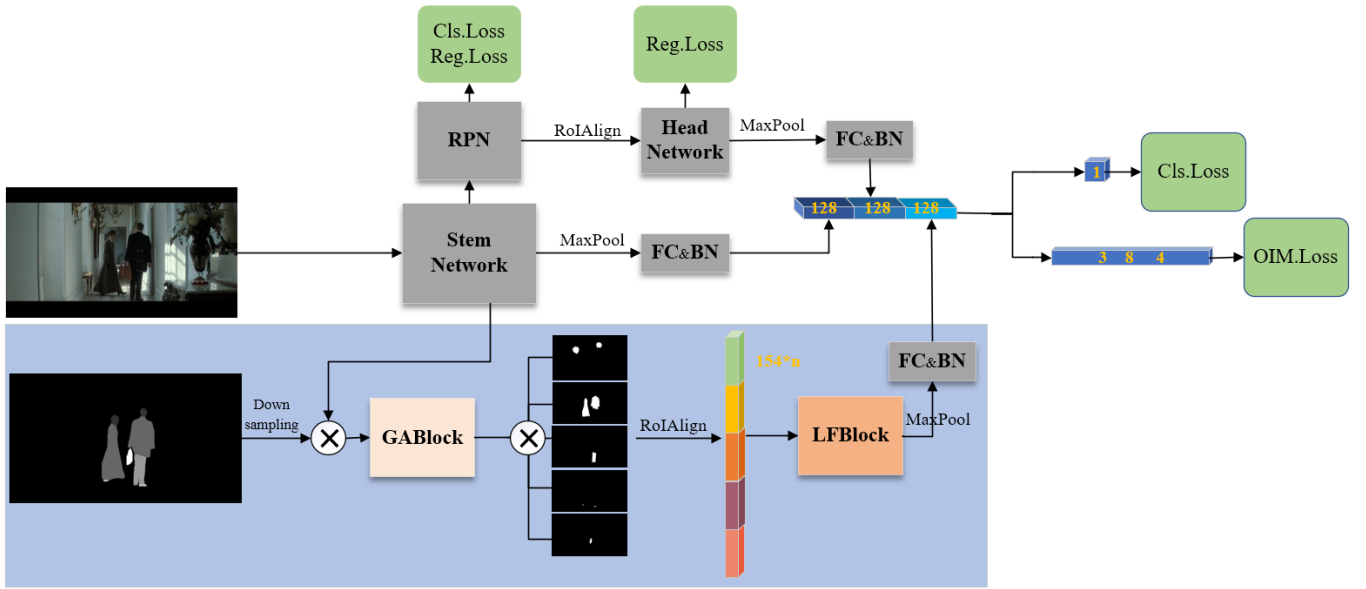

Fig. 3. Illustration of our proposed MAE. The Stem network is composed of residual blocks "Conv1" to "Conv4" of ResNet50, while the Head network is Conv5. Multi-Attribute Enhanced modules are marked blue. GABlock and LFBlock use attribute labels to learn local features, which are shown in *Section 3.2* and *Section 3.3*, respectively.

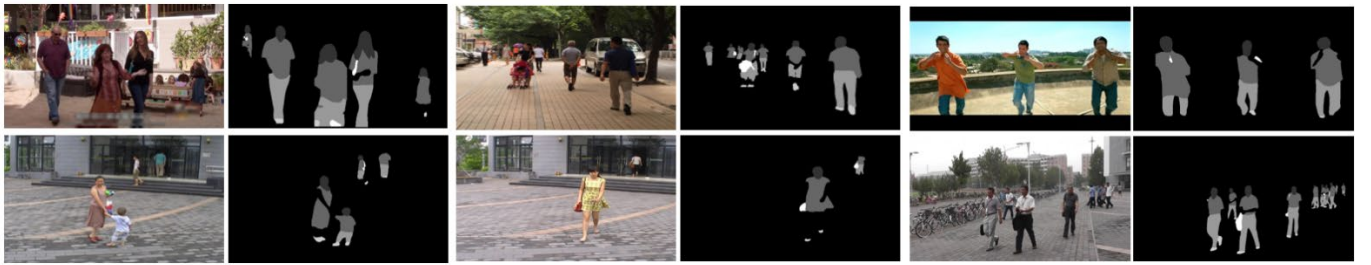

Fig. 4. Multi-attribute labels on CUHK-SYSU (the first row) and PRW (the second row) generated by HRNet. The human body does not serve as a label.

### III. MULTI-ATTRIBUTE ENHANCED NETWORK

In this part, we will show our Multi-Attribute Enhanced (MAE) model. It is worth mentioning that since the Norm-Aware Embedding [9] is very helpful in distinguishing the foreground and background feature spaces, we decide to implement our ideas on it. As shown in Fig. 3, the difference is that we use global max pool (GMP) to retain discriminative features. Before the features were decomposed in the polar coordinate system, our features were no longer generated directly by Head network and fully connected, but consisted of global features and local features.

### A. Multi-Attribute labels

In natural vision, the identification of pedestrians will integrate a variety of factors, including face, clothing, color, hairstyle, backpack, companion etc. In Re-ID, because the gallery is composed of cropped images, hand-craft horizontal stripes and pose estimation can be used for attribute segmentation or alignment. However, in Person Search, the gallery is composed of video frames that have not been cropped. Each picture contains a different number of pedestrians, and each pedestrian occupies a small proportion of the image. Therefore, the direct use of the above method will cause serious segmentation errors.

To avoid this, we propose a novel whole scene multi-attribute label. The production of the label set can be divided into three steps. Firstly, pedestrians are cropped from the gallery, and then parsed annotations are performed using HRNet [1, 4], which is retrained on multiple human parsing datasets: MHPv2 [31], ATR [32], and Viper [33]. Finally, we splice the results into the corresponding positions. So, the pedestrian in the video frame is consistent with the attributes in the label set. As shown in Fig. 4, pedestrians in CUHK-SYSU and PRW are divided into 5 categories: head, clothes on the upper body, clothes on the lower body, shoes and bags. It is worth noting that all labels do not contain human bodies.

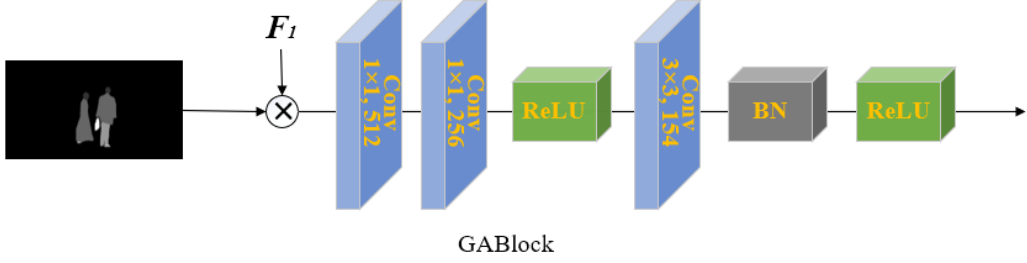

Fig. 5. The architecture of global attention Block. The module uses global labels to suppress background clutter and reduce dimension.

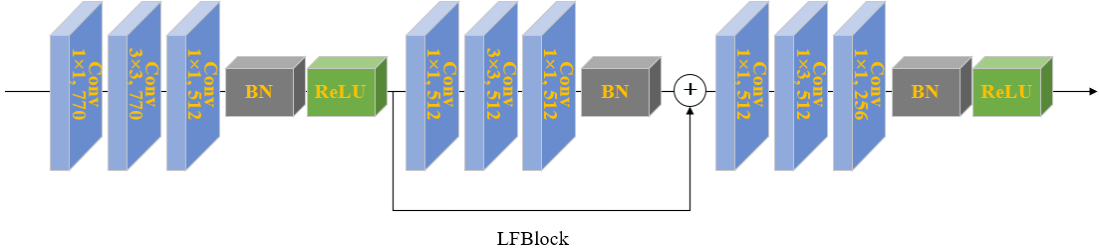

Fig. 6. The architecture of Local Fusion Block. It performs fusion learning after cascading.

In practice, the pedestrian detector used to obtain the bounding boxes from the video frame, and then cropped the corresponding pedestrian features according to the position of the bounding boxes on the feature map. In order to align the attributes with the pedestrians, we also cut the pedestrian attribute labels according to the corresponding position of the bounding boxes on the labels graph.

### B. Global Attention Block

Since the pedestrian has a small proportion in video frames, the corresponding local proportion is smaller. Meanwhile, the detector based on Fasters-RCNN [26] has a large receptive field in the process of extracting the feature layer, which will lead to the confusion between foreground and background. As shown in Fig. 5, we input images and utilize a global mask to suppress the redundant background clutter so as to better focus on the foreground details. We first calculate feature maps $F_1 \in \mathbb{R}^{1024 \times m \times n}$ obtained by the stem network and global mask $M_g \in \mathbb{N}^{1 \times m \times n}$ which is reduced to the same size by down-sampling:

$$F_2 = F_1 \odot M_g \qquad (1)$$

$$where \quad F_2 \in \mathbb{R}^{1024 \times m \times n}$$

where $\odot$ refers to element-wise multiplication, and $M_g \in \{0,1\}$.

Moreover, we aim to fuse global mask more tightly with the feature layer by the global attention block, whose structure is shown in Fig. 5, and this transformation process is defined as:

$$F_2' = \sigma(W_{g_2} \cdot W_{g_1} \cdot F_2) \qquad (2)$$

$$F_3 = \sigma(f_{BN}^g(W_{g_3} \cdot F_2')) \qquad (3)$$

$$where \quad F_3 \in \mathbb{R}^{154 \times m \times n}$$

where $W_{g_1}$, $W_{g_2}$ and $W_{g_3}$ represent the weight matrix of the corresponding convolutional layers, respectively. And $\sigma(\cdot)$ denotes the activation function of ReLU, $f_{BN}^g$ refers to Batch Normalization layer.

### C. Local Fusion Block

After the clutter is suppressed, more refined local features are extracted. As shown in Fig. 3, feature maps $F_3$ is firstly calculated with attribute labels and cascaded:

$$F_4 = [F_3 \odot M_l^1, \ldots, F_3 \odot M_l^i] \qquad (4)$$

$$where \quad F_4 \in \mathbb{R}^{770 \times m \times n}$$

$\odot$ refers to element-wise multiplication, and $M_l^i \in \{0,1\}$ denotes the i-th local attribute label. $[\cdot]$ means cascading on the channel.

We further re-weight the local feature maps with the local fusion block, which can select more discriminant attribute features under the same conditions. As shown in Fig. 6, it is defined as:

$$F_4' = \sigma(f_{BN_1}^l(W_{l_3} \cdot W_{l_2} \cdot W_{l_1} \cdot F_4)) \qquad (5)$$

$$F_4'' = f_{BN_2}^l(W_{l_6} \cdot W_{l_5} \cdot W_{l_4} \cdot F_4') + F_4' \qquad (6)$$

$$F_5 = \sigma(f_{BN_3}^l(W_{l_9} \cdot W_{l_8} \cdot W_{l_7} \cdot F_4'')) \qquad (7)$$

$$where \quad F_5 \in \mathbb{R}^{256 \times m \times n}$$

where $W_{l_1}, \ldots, W_{l_9}$ represent the weight matrix of the corresponding convolutional layers, respectively. And $\sigma(\cdot)$ denotes the activation function of ReLU, $f_{BN_1}^l, f_{BN_2}^l$ and $f_{BN_3}^l$ refer to Batch Normalization layers.

## IV. EXPERIMENTS

In this part, we introduce the implementation details of the model, and carried out a large number of experiments on two benchmark datasets to verify the effectiveness of our model.

TABLE I. EVALUATING THE EFFECTIVENESS OF MAE ON CUHK-SYSU WITH THE GALLERY SET TO 100.

| Method | Detector | | Person Search | |
|---|---|---|---|---|
| | *Recall* | *AP* | *mAP* | *rank-1* |
| OIM-re | 85.15 | 76.18 | 79.93 | 80.52 |
| NAE-re | 90.31 | 84.05 | 91.04 | 92.31 |
| OIM w/ MAE | 88.69 | 78.48 | 85.37 | 87.24 |
| MAE | 91.85 | 86.75 | 91.79 | 93.03 |

## A. Datasets and Evaluation Protocol

**CUHK-SYSU.** CUHK-SYSU [14] is derived from street snapshots and movie screenshots. It has 18,184 annotated pictures and 96,143 pedestrian bounding boxes. The training set contains 11,206 images and 5,532 query targets, while the testing set includes 6,978 images and 2,900 query targets. The training set and testing set are independent of each other without overlap.

**PRW.** PRW [20] consists entirely of outdoor video frames, including 11,816 images and 34,304 bounding boxes. The training set contains 5,134 images and 482 query targets, while the testing set contains 6,112 images and 2,057 query targets. Compared with CUHK-SYSU, it is more difficult because the probe target needs to be found in a larger gallery.

**Evaluation Protocol.** In Person Search, mAP and CMC are the most commonly used metrics to measure the search performance, and we also use them.

## B. Network and Implementation Details

Our model is implemented based on ImageNet-pretrained [34] Faster-RCNN [26]. It is mainly composed of three parts.

Firstly, Residual blocks "conv1" to "conv4" act as the stem network and freeze the parameters of the first convolution layer. It can extract rough feature maps from the original for subsequent tasks. Secondly, the head network is composed of residual block "conv5", which is used for bounding box regression and as part of the person representation. Finally, the multi-attribute enhancement module has been described in detail in *Section 3*.

As depicted in Fig. 3, the output for all RoIAlign [35] and fully connected layers is 14×14 and 128-dimensional feature maps, respectively. In the stage of training, we sample 2 images for each batch, which are resized to 900×1,500, and use gradient accumulation to average the loss of six images and then conduct a backpropagation. Our model is trained 22 and 11 epochs on CUHK-SYSU and PRW, respectively. Our device is a 16G NVIDIA Tesla V100 GPU. The initial learning rate is set to 0.003, which decreased by 10 for every 8 epochs, and is progressively warmed up during first epoch. Our optimizer uses SGD, with momentum and weight decay set to 0.9 and 5×10-4, respectively. The rest are consistent with NAE.

## C. Ablation Study

In order to prove the effectiveness of the module, the following models are tested on CUHK-SYSU with a gallery size of 100.

TABLE II. ABLATION EXPERIMENTS ON THE GLOBAL MASK.

| Method | *mAP* | ▲ | *rank-1* | ▲ |
|---|---|---|---|---|
| MAE w/o global mask | 91.47 | +0.32 | 92.24 | +0.79 |
| MAE | 91.79 | | 93.03 | |

TABLE III. COMPARISON WITH STATE-OF-THE-ARTS ON CUHK-SYSU (GALLERY SIZE IS SET TO 100) AND PRW (GALLERY SIZE IS SET TO 6,112).

| Method | CUHK-SYSU | | PRW | |
|---|---|---|---|---|
| | *mAP* | *rank-1* | *mAP* | *rank-1* |
| OIM [14] | 75.5 | 78.7 | 21.3 | 49.9 |
| OIM-re [14] | 79.9 | 80.5 | 25.5 | 72.4 |
| IAN [3] | 76.3 | 80.1 | 23.0 | 61.9 |
| NPSM [7] | 77.9 | 81.2 | 24.2 | 53.1 |
| MGTS [2] | 83.0 | 83.7 | 32.6 | 72.1 |
| GRAPH [18] | 84.1 | 86.5 | 33.4 | 73.6 |
| QEEPS [15] | 88.9 | 89.1 | 37.1 | 76.7 |
| APNet [13] | 88.9 | 89.3 | 41.9 | 81.4 |
| BINet [8] | 90.0 | 90.7 | **45.3** | **81.7** |
| NAE [9] | 91.5 | 92.4 | 43.3 | 80.9 |
| NAE-re [9] | 91.0 | 92.3 | 36.5 | 76.5 |
| OIM-re w/ MAE (our) | 85.4 | 87.2 | 30.9 | 73.9 |
| MAE (our) | **91.8** | **93.0** | 40.1 | 79.7 |

**NAE-re:** Due to the limitation of computing resources, we could not reproduce the results in NAE [9]. In order to make the results comparable, we reproduced them on the same strategy and device. The main change is to set the batch from 5 to 2 and use the same cumulative optimizations and learning rates described in *Section 4.1*. It can be seen that our results are close to those of the original papers.

**OIM-re:** We re-implemented OIM [14], and the experimental results are slightly better than the original.

**MAE:** We introduced the multi-attribute enhancement model in *Section 3*.

**OIM w/ MAE:** Add MAE module to OIM.

Since the performance of search is significantly affected by the bounding boxes, the quality of the pedestrian detector is also taken into account. Both Recall and Average Precision (AP) are used as metrics.

**Effectiveness of the MAE.** As shown in Tab. 1, the detector of MAE reaches 91.85% and 86.75% w.r.t. Recall and AP exceed NAE-re by 1.54% and 2.70%, respectively. The MAE reaches 91.79% and 93.03% w.r.t. mAP and rank-1, surpassing NAE-re by 0.75% and 0.72%, respectively. The situation also occurs in experiments based on OIM, and the corresponding performance improvement is more significant. The detector of OIM w/ MAE exceeds OIM-re by 3.54% and 2.30% on Recall and AP, respectively. The OIM w/ MAE surpasses OIM-re by 5.44% and 6.72% on mAP and rank-1, respectively. These indicate that our MAE helps to generate high-quality bounding boxes, and more importantly, improves

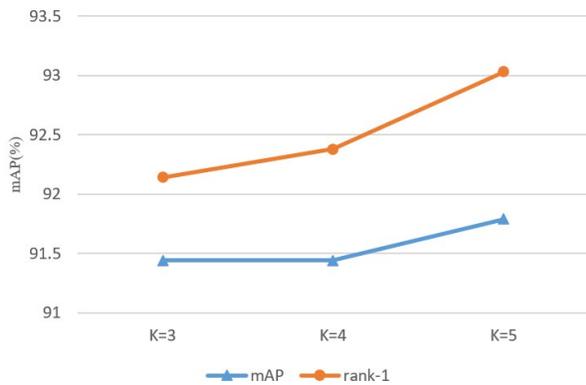

Fig. 7. MAE performance on CUHK-SYSU for different attribute partitioning K.

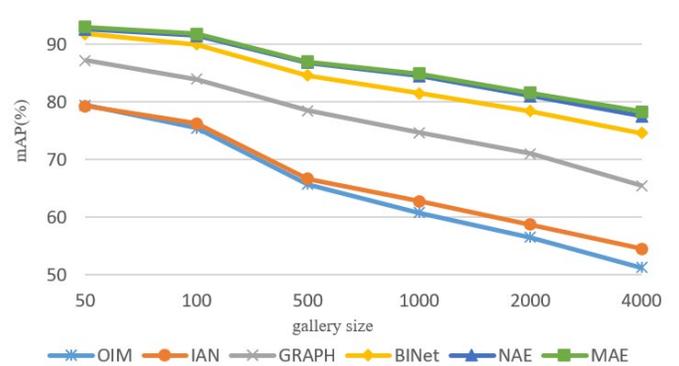

Fig. 8. Evaluation on CUHK-SYSU with different gallery sizes.

search performance.

**Global mask helps suppress clutter.** As mentioned in *Section 3.2*, because the pedestrian has a small proportion in images, the pedestrian local proportion is smaller. Meanwhile, the detector of MAE has a large receptive field in the process of extracting the feature layer, which will lead to the confusion between foreground and background. It is an effective method to use global masks to suppress background clutter. This is supported by the data in Tab. 2, where the return on mAP and rank-1 are 0.32% and 0.79%, respectively.

**How to divide attributes?** In order to explore the impact of attribute division on model performance, we divide the five attributes introduced in *Section 3.1* into three categories. K=3: head, all-clothes (clothes on the upper body, clothes on the lower body and shoes are merged.), and bags. K=4: head, clothes on the upper body, lower-clothes (clothes on the lower body and shoes are merged.), and bags. K=5: no labels are merged. As shown in Fig. 7, as the attribute partitioning becomes more detailed, the performance improves as well as. Compared with the ambiguity of global features, this may be due to the improvement of K that leads to learning more representative attributes of pedestrians.

### D. Comparison to the State-of-the-arts

**Comparison on CUHK-SYSU.** As shown in Tab. 3, the MAE reaches 91.8% and 93.0% w.r.t. mAP and rank-1. The performance of MAE is the best among end-to-end methods, including the robust models APNet [13] and BINet [8]. Note that their forward process consumes a lot of resources, especially the Siamese network to deal with human patches in BINet. In contrast, we have only one forward transfer, and the attribute label operates directly with the feature map so as to reduce the overhead of computational resources. Meanwhile, it is worth mentioning that our model still surpasses NAE with fewer computing resources and suboptimal hyperparameters. With the addition of our module, OIM-re improved by 5.5% and 6.3% w.r.t. mAP and rank-1. The above shows that our module is effective and reliable.

To evaluate the stability of our algorithm, we compare it with other end-to-end methods in larger search scopes. As shown in Fig. 8, as the size of the gallery increases from 50 to 4,000, and introduces more interference, the performance of all methods has decreased to varying degrees. However, MAE achieves the best performance under the same size. In addition, the mAP of MAE drops from 93.0% to 78.3% when the size changes from 50 to 4,000, which shows a more moderate downward trend compared with other methods and proves the robustness of our model.

**Comparison on PRW.** We further evaluate our model on the PRW, and the results are shown in Tab. 3. Our MAE improve by 3.6% and 3.2% w.r.t. mAP and rank-1 compared to NAE-re. The module we add to the OIM-re also gets a big performance boost. We notice that the results of APNet [13] and BINet [8] are better than ours. APNet estimates and refines pedestrians in the bounding boxes, extracts the human body from bounding boxes, discards features of occlusion or noise area, and then performs alignment matching. BINet uses Siamese network to add an instance-aware branch. Besides video frames, it can also take human patches as input to help suppress redundant context information outside the bounding boxes. Therefore, the methods used by APNet and BINet are easier to identify blocked pedestrians. However, CUHK-SYSU does not include bounding boxes with partial bodies, and PRW annotates some misaligned bounding boxes in both queries and galleries. So, the performance improvement of APNet and BINet on PRW is more significant than that on CUHK-SYSU. And we did not specifically optimize for this problem so that our performance was lower than APNet and BINet on PRW.

In addition, compared with CUHK-SYSU, the larger image size leads to a smaller proportion of pedestrian scales in PRW. However, the size of the receptive field of the model is the same, so the foreground attribute features are more susceptible to interference from background clutter. Meanwhile, it can be found that the performance of NAE on PRW is sensitive to hyperparameters and computing resources. Compared with NAE, the performance of NAE-re decreased by 6.8% and 4.4% w.r.t. mAP and rank-1. Therefore, our results are valid and interpretable, and we believe that the above problems can be alleviated by using the Siamese network mentioned by BINet [8], in which one is input video frame and the other is input pedestrian attribute.

### E. Visual comparison

In Fig. 9, we also show some qualitative results of our MAE and the baseline NAE. As we can see from Fig. 9, in (a) ~ (c), the target in the query image and the gallery have the same viewpoint, e.g. the camera is on the front or back of the target. Both our method and NAE obtained correct results. However, as the viewpoint of camera changed or the target became smaller, NAE misjudged and our method still retrieved the correct target. For example, in (d) ~ (f), query image has a different viewpoint from the gallery, e.g. the camera is on the front of the target in the query image, while the camera is on the back in the gallery. It also happens in (g) ~ (i), the camera away from the pedestrian makes the query target or pedestrian smaller in the gallery. In general, our model is more robust when the query target changes or crosses cameras, and can provide more discriminative features according to the attributes.

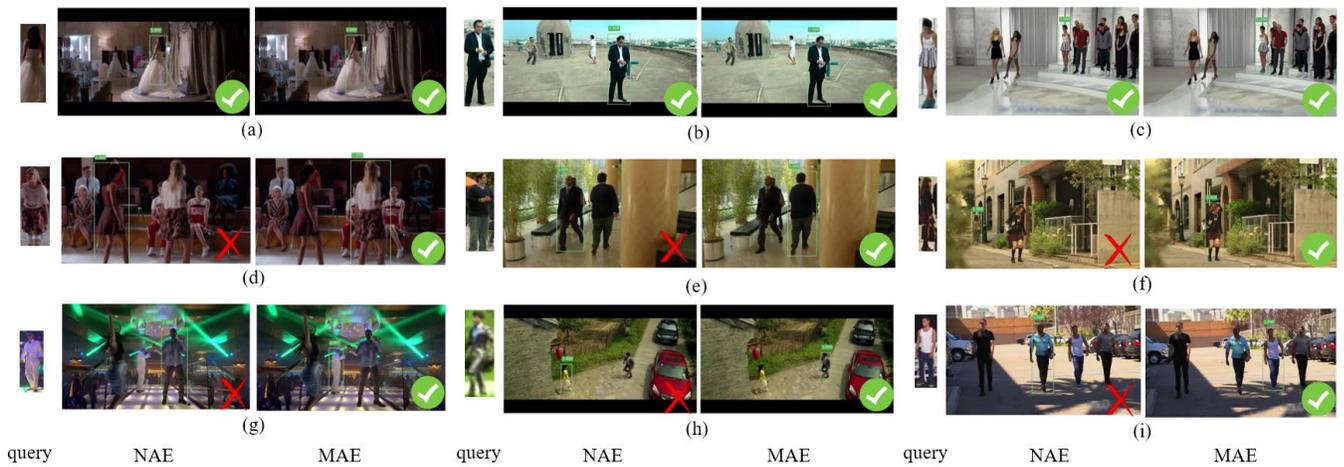

Fig. 9. Rank-1 search results for several samples. For each query image, we show the rank-1 match given by NAE and our MAE. Green/red markers indicate correct/false results, respectively. (a) ~ (c): query image has the same viewpoint as the gallery. (d) ~ (f): query image has a different viewpoint from the gallery. (g) ~ (i): the target in the query image and gallery is smaller.

## V. CONCLUSION

In this paper, we propose a novel end-to-end network, named Multi-Attribute Enhanced. Different from the previous methods, we no longer use global features alone for matching, and our model can use attribute labels to learn the attribute representation for matching task. This method mainly conducts joint learning of attribute labels and coarse feature maps, which are obtained from ResNet50, and then applies local and global representations to subsequent tasks. Compared with Pose Estimation and other methods to achieve attribute alignment, our Multi-attribute labels not only simplifies the model structure but is also computationally resource-friendly. We have conducted extensive experiments to prove that our module can significantly improve the search performance.

Our model requires video frames to obtain attribute labels in advance. Although the generation of these labels is convenient and efficient, the overall steps are not simple enough. Therefore, in the future work, we will explore the direct learning of pedestrian local representation in the joint framework, so as to the process more concise.

## ACKNOWLEDGMENT

This work was supported in part by National Natural Science Foundation of China (61501198), the Fundamental Research Funds for the Central Universities (CCNU20TS028) and Teaching research project of CCNU (202013).